\documentclass[sn-mathphys]{sn-jnl}
\jyear{2022}%
\usepackage{hyperref}
\usepackage{multirow}

\newcommand{\ti}{\textit}

\begin{document}
\title[Performance variability in polyp detection]{Sources of performance variability in deep learning-based polyp detection}

\author*[1]{\fnm{T.\,N.} \sur{Tran}}\email{t.tran@dkfz-heidelberg.de}

\author[1]{\fnm{T.\,J.} \sur{Adler}}
\author[1]{\fnm{A.} \sur{Yamlahi}}
\author[1]{\fnm{E.} \sur{Christodoulou}}
\author[1]{\fnm{P.} \sur{Godau }}
\author[1]{\fnm{A.} \sur{Reinke }}
\author[1]{\fnm{M.\,D.} \sur{Tizabi}}
\author[2]{\fnm{P.} \sur{Sauer}}
\author[3]{\fnm{T.} \sur{Persicke}}
\author[3,4]{\fnm{J.\,G.} \sur{Albert}}
\author[1]{\fnm{L.} \sur{Maier-Hein}}
\affil[1]{\orgdiv{Div. Intelligent Medical Systems}, \orgname{DKFZ}, \orgaddress{\country{Germany}}}
\affil[2]{\orgdiv{Interdisciplinary Endoscopy Center (IEZ)}, \orgname{University Hospital Heidelberg},\orgaddress{ \country{Germany}}}
\affil[3]{\orgdiv{Department of Gastroenterology, Hepatology and Endocrinology}, \orgname{Robert-Bosch Hospital (RBK)}, \country{Germany}}
\affil[4]{\orgdiv{Clinic for General Internal Medicine, Gastroenterology, Hepatology and Infectiology, Pneumology}, \orgname{Klinikum Stuttgart}, \orgaddress{ \country{Germany}}}

\abstract{Validation metrics are a key prerequisite for the reliable tracking of scientific progress and for deciding on the potential clinical translation of methods. While recent initiatives aim to develop comprehensive theoretical frameworks for understanding metric-related pitfalls in image analysis problems, there is a lack of experimental evidence on the concrete effects of common and rare pitfalls on specific applications. We address this gap in the literature in the context of colon cancer screening. Our contribution is twofold. Firstly, we present the winning solution of the Endoscopy Computer Vision Challenge (EndoCV) on colon cancer detection, conducted in conjunction with the IEEE International Symposium on Biomedical Imaging (ISBI) 2022. Secondly, we demonstrate the sensitivity of commonly used metrics to a range of hyperparameters as well as the consequences of poor metric choices. Based on comprehensive validation studies performed with patient data from six clinical centers, we found all commonly applied object detection metrics to be subject to high inter-center variability. Furthermore, our results clearly demonstrate that the adaptation of standard hyperparameters used in the computer vision community does not generally lead to the clinically most plausible results. Finally, we present localization criteria that correspond well to clinical relevance. Our work could be a first step towards reconsidering common validation strategies in automatic colon cancer screening applications.
}

\keywords{Validation, Evaluation, Metrics, Object detection, Surgical data science, Variability}

\maketitle

\section{Introduction}\label{sec:introduction}
Colorectal cancer is one of the most common cancer types, ranking second in females and third in males \cite{haggar2009colorectal}. By detecting and subsequently resecting neoplastic polyps during screening colonoscopy, the risk of developing the disease can be reduced significantly. Research focuses on developing deep learning (DL) solutions for automated detection of polyps in colonoscopy videos \cite{fitting2022video,Ali2022EndoCV21,Ali2022EndoCV22,Bernal2021giana21,Bernal2021giana1718}. However, to date, the metrics with which algorithms are validated receive far too little attention. These metrics are not only important for measuring scientific progress, but also for gauging a method’s potential for clinical translation. While previous work has highlighted general metric pitfalls in the broader context of classification, segmentation and detection \cite{Reinke2021CommonLO}, we are not aware of any prior studies systematically analyzing common metrics in the context of polyp detection. Our underlying hypothesis was that reported performance values in polyp detection methods are largely misleading as they are sensitive to many validation design choices including (1) the choice of test set and (2) the chosen metric configurations (e.g. threshold for the localization criteria).
Our contribution is twofold: Firstly, we present the winning solution of the Endoscopy Computer Vision Challenge (EndoCV) on colon cancer detection, conducted in conjunction with the IEEE International Symposium on Biomedical Imaging (ISBI) 2022. Secondly, based on publicly available challenge data, we demonstrate the sensitivity of commonly used metrics to a range of hyperparameters as well as the consequences of poor metric choices. 

\section{Methods}\label{sec:methods}
Here, we present the winning method of the EndoCV challenge on colon cancer detection, conducted in conjunction with ISBI 2022 (Sec. \ref{sec:ObjectDAL}),  and revisit common detection metrics including their hyperparameters (Sec. \ref{sec:ObjectDetectMetric}).

\subsection{Object detection algorithm}\label{sec:ObjectDAL}
\begin{figure}[h]
\begin{minipage}[b]{1.0\linewidth}
  \centering
  \centerline{\includegraphics[width=6.5cm]{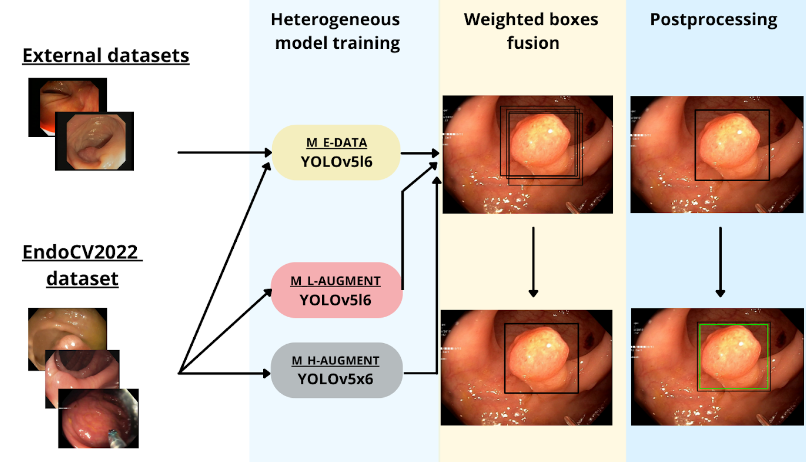}}
\end{minipage}
\caption{Winning submission of the Endoscopy computer vision challenge (EndoCV) on colon cancer detection. An ensemble of six YOLOv5-based models, each trained with different data and/or augmentation strategies, predicts a set of bounding box candidates. These are merged using weighted boxes fusion and postprocessed to yield the final prediction.
}
\label{fig:architecture_ph}
\end{figure}

We base our study on a state-of-the-art detection method, namely the winning entry \cite{yamlahi2022heterogeneous} of the EndoCV 2022 polyp detection challenge \cite{Ali2022EndoCV22}.\\
\noindent\textbf{Method overview:} The method is illustrated in Fig. \ref{fig:architecture_ph}. A heterogeneous ensemble of YOLOv5-based models was trained. To this end, we split the training data into subsets. To avoid data leakage, the split was performed along each sequence ID. Originally, we created four folds for stratified four-fold cross-validation, but the final models were trained on only two of the four folds due to training and inference time restrictions. Furthermore, we trained each model either with light augmentation on EndoCV data only, heavy augmentation on EndoCV data only, or light augmentation on EndoCV data and external data (see \cite{yamlahi2022heterogeneous} for details). Overall, this led to six ensemble members. The individual member predictions were merged using the weighted boxes fusion algorithm. As we observed a tendency towards oversegmentation, we added a postprocessing step to shrink the bounding boxes. \\
\noindent\textbf{Implementation details:} The models were trained for 20 epochs using a stochastic gradient descent optimizer, a learning rate of 0.1, and a \ti{complete Intersection over Union (CIoU)} loss. The non-maximum suppression algorithm (NMS) was applied to each ensemble member individually with an \textit{Intersection over Union (IoU)} threshold of 0.5. For the weighted boxes fusion algorithm hyperparameters, we chose an \ti{IoU} threshold of 0.5, a skip box threshold of 0.02, and all models were weighted equally. During postprocessing, we shrank all bounding boxes with a confidence score higher than 0.4 by 2\% of their size. We evaluated the ensemble a single time on our test data set.

\subsection{Object detection metrics}\label{sec:ObjectDetectMetric}
\begin{figure}[h]
\begin{minipage}[]{1.0\linewidth}
  \centering
  \centerline{\includegraphics[width=\textwidth]{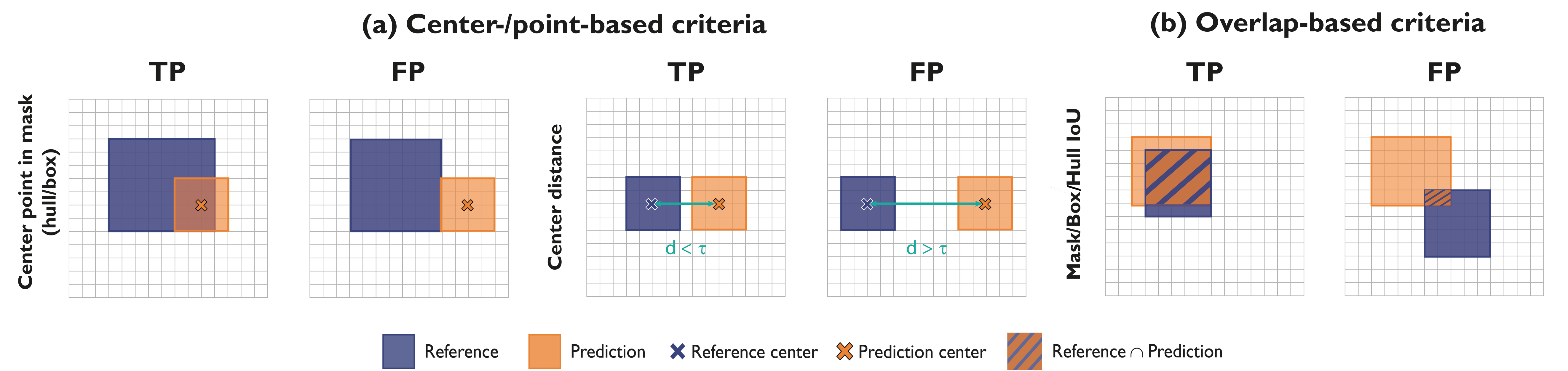}}
\end{minipage}
\caption{Localization criteria can be \textit{point-based} or \textit{overlap-based} depending on whether the user is mainly interested in the position or in the rough outline of an object. \ti{Point in mask} returns a true positive (TP) if the center point of the predicted bounding box lies within the respective reference mask. The reference can be the segmentation mask, convex hull or bounding box. Center distance criterion determines a TP if the distance \(d\) between prediction and reference centers is within a range $\tau$. For overlap-based criteria, the result is a TP if the overlap lies above a certain threshold. Depending on whether the \ti{Intersection over Union (IoU)} is computed for a reference mask or an approximating bounding box, we refer to it as \textit{Mask} or \textit{Box IoU}.}
\label{fig:criterion}
\end{figure}

\begin{figure}[h]
\begin{minipage}[]{1.0\linewidth}
  \centering
  \centerline{\includegraphics[width=9.5cm]{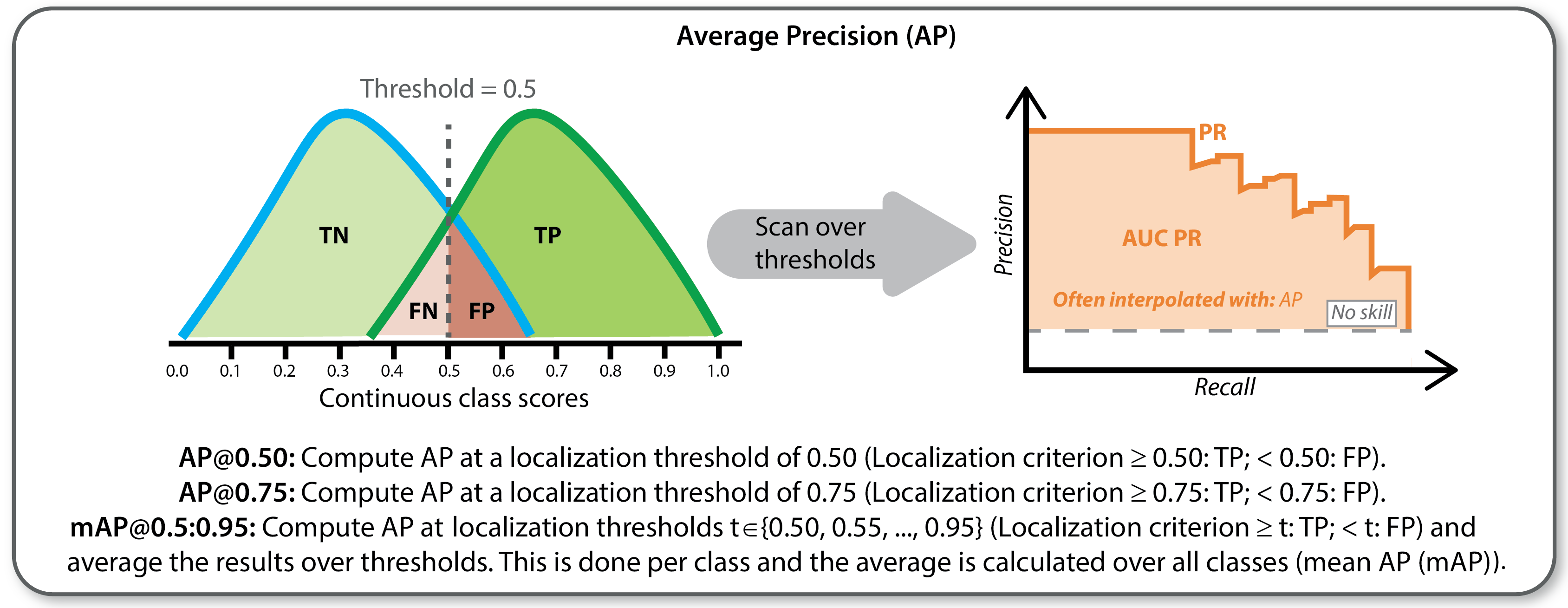}}
\end{minipage}
\caption{Pictorial representation of the \textit{Average Precision (AP)} metric.}
\label{fig:metrics3}
\end{figure}

Three metric-related design decisions are important when assessing performance of object detection algorithms  \cite{MaierHein2022Metrics}:\\
\ti{(1) Localization criterion:} The localization criterion determines whether a predicted object spatially corresponds to one of the reference objects and vice versa by measuring the spatial similarity between prediction (represented by a bounding box, pixel mask, center point or similar) and reference object. It defines whether the prediction hit/detected (true positive) or missed (false positive) the reference. Any reference object not detected by the algorithm is defined as false negative. The localization criteria that were applied in this work comprise two groups, namely the point-based criteria and the overlap-based criteria (Fig. \ref{fig:criterion}).\\
\ti{(2) Assignment strategy:} As applying the localization criterion might lead to ambiguous matchings, such as two predictions being assigned to the same reference object, an assignment strategy needs to be chosen that determines how potential ambiguities are resolved. As multiple polyps in the same image are rather rare, an assignment strategy is not as relevant as in other applications. With respect to the metric configuration, we therefore  focus on the localization criterion and the classification metrics.\\  
\ti{(3) Classification metric:} Based on the choice of localization criterion and assignment strategy, standard classification metrics can be computed at object level~\cite{Reinke2021CommonLO}. The most popular multi-threshold metric in object detection is \ti{Average Precision (AP)} (Fig.~\ref{fig:metrics3}). 
 
 As a foundation of this work, we determined common metrics in object detection challenges, along with their respective localization criterion and classification metric (Tab. \ref{tab:challengemetric}).

\begin{table}[]
\begin{center}
\begin{minipage}{0.99\linewidth}
\begin{tabular}{llll}
\toprule
 &
  Localization criterion &
  Classification metric \\ 
 \cmidrule(l){2-3}
EndoVis 2015 \cite{Bernal2017ComparativeVO} &  \ti{\textbf{Point inside mask}} &  {\color[HTML]{212529} \textbf{\begin{tabular}[c]{@{}l@{}}\textit{PPV}\\ \textit{Sensitivity }\\ \textit{Overall/Average F1-score}\end{tabular}}} \\ 
  \midrule
\begin{tabular}[c]{@{}l@{}}GIANA 2017 \cite{Bernal2021giana1718}\\ GIANA 2018 \cite{Bernal2021giana1718}\\ GIANA 2021 \cite{Bernal2021giana21}\end{tabular} &
  \ti{\textbf{Point inside mask}} &
  {\color[HTML]{212529} \textbf{\begin{tabular}[c]{@{}l@{}}\ti{PPV}\\ \ti{Sensitivity }\\ \ti{F1/F2-score}\\ Custom metrics\end{tabular}}} \\ \midrule
\begin{tabular}[c]{@{}l@{}}EndoCV 2021 \cite{Ali2022EndoCV21}\\ EndoCV 2022 \cite{Ali2022EndoCV22}\end{tabular} &
  \textbf{\begin{tabular}[c]{@{}l@{}}\textit{Box IoU} \end{tabular}} &
  \begin{tabular}[c]{@{}l@{}}\textbf{\textit{AP@IoU=0.5/0.75/{[}0.50:0.05:0.95{]}}} \\ \textbf{\textit{AP} across 3 scales of polyp size} \\ \textbf{Mean of 4 \textit{AP}s}\end{tabular} \\ 
\botrule
\end{tabular}
\end{minipage}
\end{center}
\caption{Common design choices for validation of polyp detection methods according to international competitions. \ti{PPV: Positive Predictive Value, AP: Average Precision.}
}
\label{tab:challengemetric}
\end{table}


\section{Experiments and Results}\label{sec:ExpandRes}
In this section, we investigate the sensitivity of popular classification metrics to the test set composition (Sec. \ref{sec:EffectTest}) and the localization criterion (Sec. \ref{sec:EffectMetConfig}). We further assess the clinical value of commonly used metric configurations (Sec. \ref{sec:ReflectionDOM}).

\subsection{Effect of test set}\label{sec:EffectTest}
In the following, we quantitatively assess the performance variability resulting from the chosen test set, specifically from the target domain (i.e. the clinical validation center) and the distribution of polyp size.\\

\begin{figure}[h]
\begin{minipage}[]{1.0\linewidth}
  \centering
  \centerline{\includegraphics[width=8.5cm]{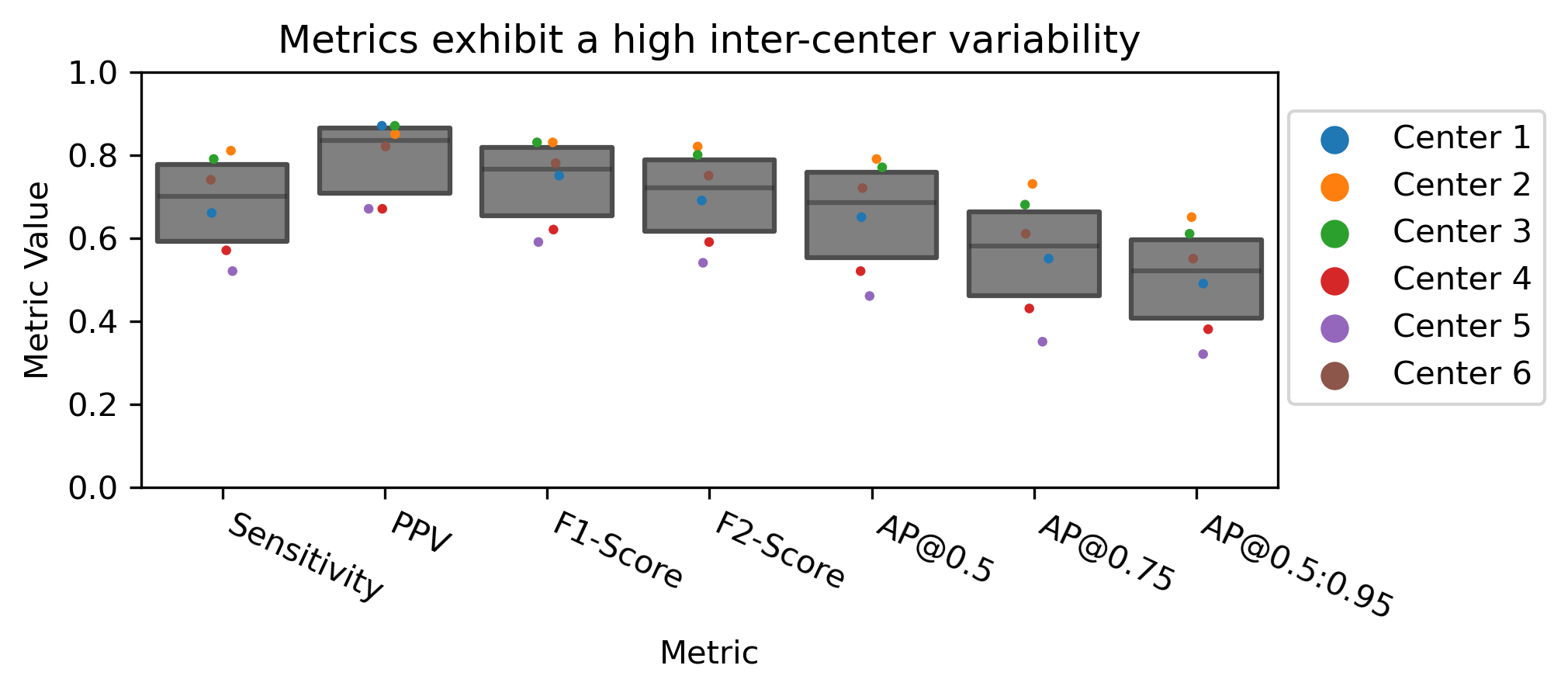}}
\end{minipage}
\caption{ Performance variability resulting from the chosen validation center. All commonly used classification metrics (cf. Tab. \ref{tab:challengemetric}) show a substantial sensitivity to the center. The dot-and-box plots contain aggregated values per center.}
\label{fig:center_detection_metric_variability}
\end{figure}

\noindent\textbf{Sensitivity to center:} To show the variability of performance resulting from different test sets, we used data from six validation centers \cite{ali2021polypgen}. Fig. \ref{fig:center_detection_metric_variability} shows the performance of our object detection method (Sec. \ref{sec:ObjectDAL}) according to commonly used metrics. These exhibit high variability between centers. For example, the \textit{AP} ranges from [0.38, 0.65], which is notable, given that the \ti{AP} of the top three submissions for EndoCV 2022 ranged from [0.12, 0.33].\\

\noindent\textbf{Sensitivity to polyp size:} Using the polyp size definitions introduced by the EndoCV 2021 challenge \cite{Ali2022EndoCV21}, we further calculated the \ti{AP} scores from all six validation centers, stratified by polyp size (Tab. \ref{tab:polypsizes}). A high variability can be observed, indicating that algorithm performance is highly affected by the distribution of polyp sizes.

\begin{table}[h]
\begin{center}
\begin{tabular}{lcccccccc}
\toprule
Metric &  \multicolumn{3}{@{}c@{}}{\ti{AP@0.5}} & \multicolumn{3}{@{}c@{}}{\ti{AP@{[}0.5:0.95{]}}} & \multirow{2}[3]{*}{n} & \multirow{2}[3]{*}{\(\phi\)} \\
\cmidrule(r){2-4}\cmidrule(r){5-7}
Polyp size & small & medium & large & small & medium & large & & \\
\midrule
Center 1 & 0.24 & 0.48 & 0.73 & 0.14 & 0.33 & 0.55 & 256 & 99\% \\
Center 2 & 0.10 & 0.58 & 0.85 & 0.04 & 0.42 & 0.70 & 301 & 90\%\\
Center 3 & 0.16 & 0.64 & 0.91 & 0.10 & 0.47 & 0.72 & 457 & 99\% \\
Center 4 & 0.14 & 0.33 & 0.59 & 0.06 & 0.27 & 0.43 & 227 & 64\% \\
Center 5 & 0.19 & 0.39 & 0.59 & 0.12 & 0.26 & 0.42 & 208 & 99\%\\
Center 6 & 0.00 & 0.52 & 0.89 & 0.00 & 0.39 & 0.67 & 88 & 94\%\\
\midrule
\textbf{\begin{tabular}[c]{@{}l@{}}All Centers\\ (SD)*\end{tabular}} & \begin{tabular}[c]{@{}c@{}}0.13\\ (0.08)\end{tabular} & \begin{tabular}[c]{@{}c@{}}0.49\\ (0.12)\end{tabular} & \begin{tabular}[c]{@{}c@{}}0.76\\ (0.15)\end{tabular} &
\begin{tabular}[c]{@{}c@{}}0.08\\ (0.05)\end{tabular} & \begin{tabular}[c]{@{}c@{}}0.36 \\ (0.08)\end{tabular} & \begin{tabular}[c]{@{}c@{}}0.58 \\ (0.12)\end{tabular} & 1537 & 91\%\\
\botrule
\end{tabular}
\end{center}
\caption{\ti{Average Precision (AP)} stratified by polyp size. The results are shown for a fixed \ti{Intersection over Union (IoU)} threshold of 0.5 (left) as well as for a range of thresholds following the COCO benchmark evaluation standard \cite{lin2014microsoft} (center).  We provide additional information on the number of frames (n) and polyp prevalence ($\phi$) per center (right).
*SD: standard deviation
}
\label{tab:polypsizes}
\end{table}
To further evaluate how the \ti{IoU} values relate to polyp size and polyp type and simultaneously account for the hierarchical structure of the data set, we fit a linear mixed effects model (R version 4.1.3, package lme4). In this model, polyp size (small, medium, or large) and polyp type (flat or protruded) were fixed effects, while data center, patient identifier (ID), and image ID were random effects. The results suggest that there are strong effects of polyp type and polyp size on the  \ti{IoU} values. In particular when the polyp is of a protruded as opposed to a flat type, the values of \ti{IoU} are on average higher by a difference of 0.08 (conditional that the other predictors remain constant). When the polyp is of a medium or small size compared to a large size, the \ti{IoU} values are lower by a difference of 0.05 and 0.28, respectively (conditional that the remaining predictors remain constant). 

\subsection{Effect of metric configuration}\label{sec:EffectMetConfig}
\begin{figure*}[t]
\begin{center}
   {\includegraphics[width=\textwidth]{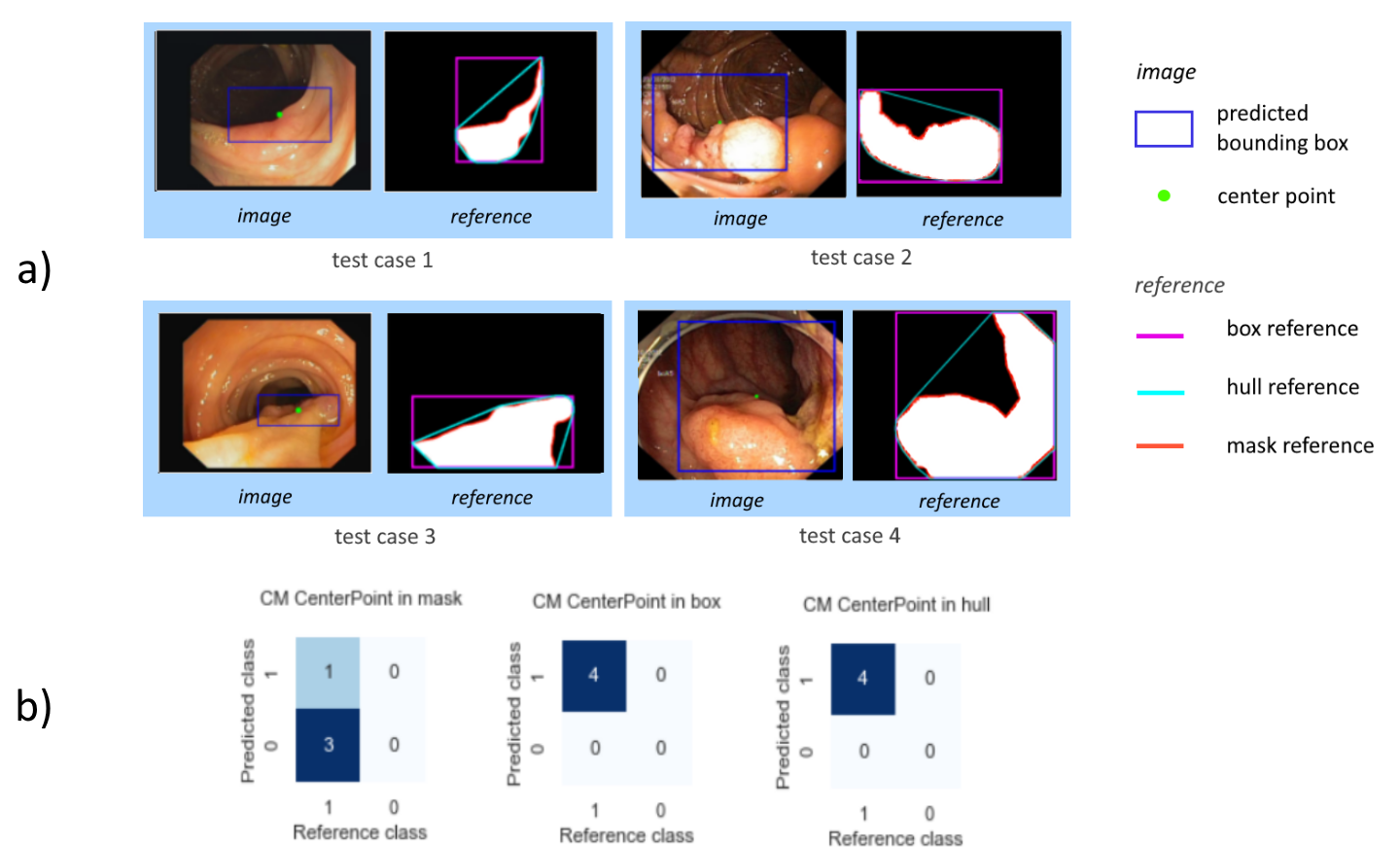}}\\
\end{center}
\caption{Effect of the reference shape in point-based localization criteria (a) on the confusion matrix (CM) (b). In the case of non-convex polyps, \textit{Mask IoU} leads to substantially more false positives. 
}
\label{fig:hull_box_mask_legend3.png}
\end{figure*}

\begin{figure*}[t]
  \centering
   {\includegraphics[width=\textwidth]{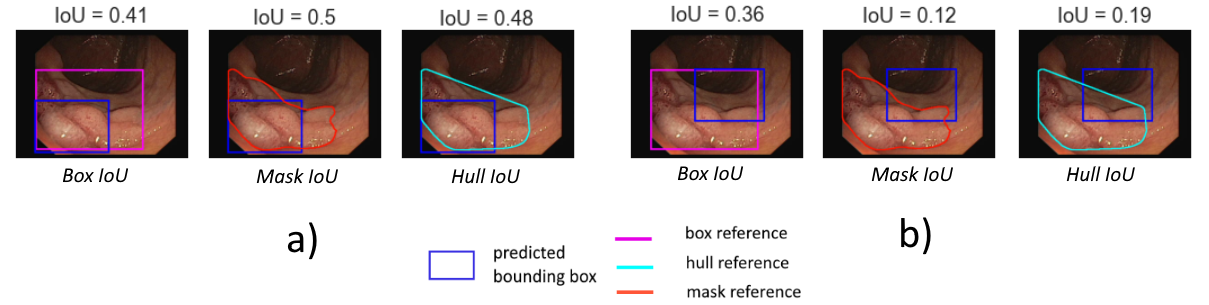}}\\
\caption{Effect of the reference shape (here: reference mask or its bounding box or convex hull) in boundary-based localization criteria. For two different (blue) predictions (a) and (b) the \textit{Intersection over Union (IoU)} results are shown. These vary substantially in the case of the inferior prediction (b). 
}
\label{fig:iou.png}
\end{figure*}

\begin{figure*}[t]
  \centering
  {\includegraphics[width=\textwidth]{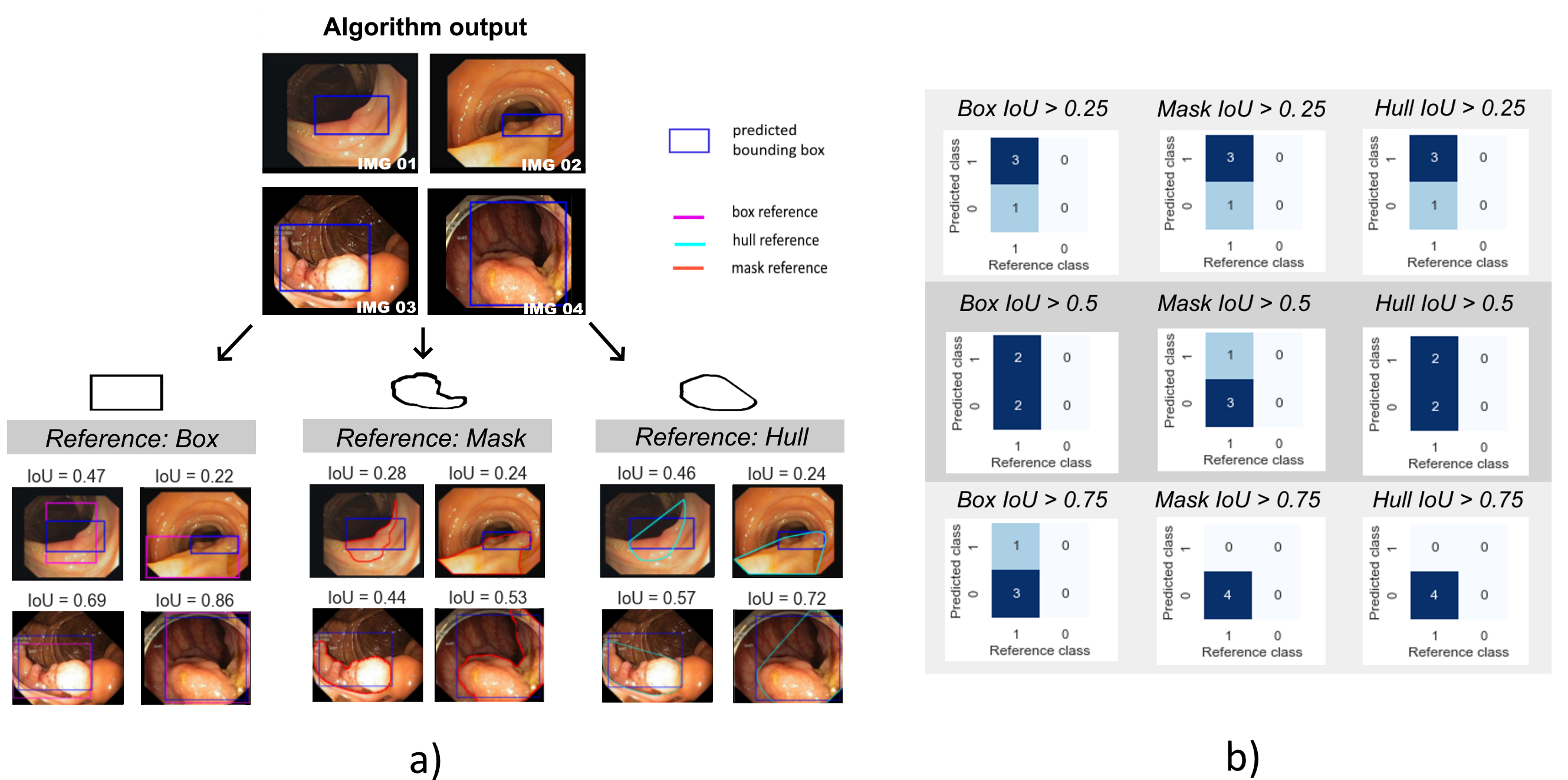}}\\
\caption{Effect of \textit{Intersection over Union (IoU)} threshold on the confusion matrix for three different overlap-based localization criteria. The same predictions produce substantially different confusion matrices for commonly used thresholds 0.5 and 0.75. 
}
\label{fig:overlap_scheme3}
\end{figure*}

In the case of polyp detection, the goal of high sensitivity (not missing a polyp) is an indispensable priority. We therefore assess the effect of design choices related to the localization criterion on the decision whether a prediction is determined to be a true or false positive. 
Figures \ref{fig:hull_box_mask_legend3.png} and \ref{fig:iou.png} showcase the effect of the reference shape in point-based and overlap-based localization criteria, respectively, while Fig. \ref{fig:overlap_scheme3} demonstrates the sensitivity of overlap-based criteria to different localization thresholds. 
In the following, we provide experimental evidence for the showcased phenomena.\\

\begin{figure}[]
\begin{center}
  \includegraphics[width=0.9\linewidth]{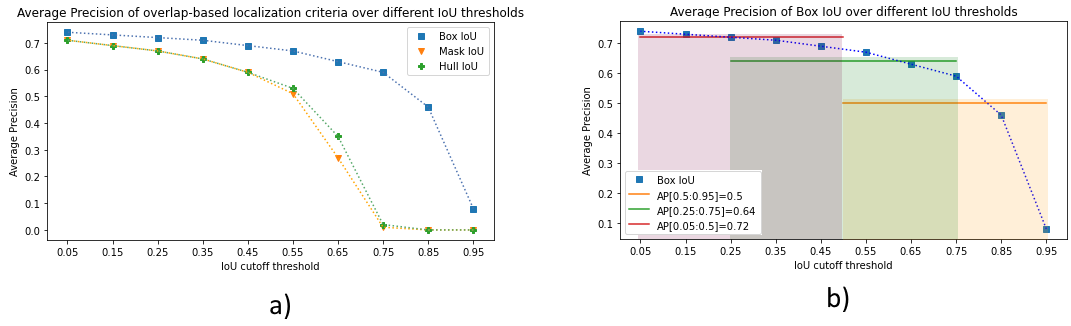}
\end{center}
\caption{(a) Effect of different localization criteria on the most common object detection metric \ti{Average Precision (AP)}. Three common overlap-based criteria using different references (box, mask and hull) are plotted as a function of the \ti{Intersection over Union (IoU)} cutoff threshold in the range [0.05:0.95]. \textit{Box IoU} scores are higher across all thresholds, while \textit{Mask IoU} and \textit{Hull IoU} do not differ substantially. (b) \textit{Average Precision (AP)} with \textit{Box Intersection over Union (IoU)} threshold for three different ranges of \textit{IoU} thresholds. Note that the range [0.5:0.95] (orange) is the most common one in the computer vision community.}
\label{fig:AP_iou2}
\end{figure}

\noindent\textbf{Sensitivity of the AP to the specific choice of overlap-based localization criterion:} In this experiment, we investigated the \ti{AP} scores using \textit{Box IoU, Mask IoU} and \textit{Hull IoU} criteria over a range of \textit{IoU} thresholds [0.05:0.95]. The resulting curves are shown in Fig. \ref{fig:AP_iou2}a). We observe that the \textit{Mask IoU} and \textit{Hull IoU}-based \textit{AP} scores are very similar; conversely,  using \textit{Box IoU} yielded overall higher \textit{AP}, even at lower thresholds.\\

\noindent\textbf{Sensitivity of the \textit{AP} to the \textit{IoU} range:} We investigated the \textit{AP} scores, using \textit{Box IoU} as a criterion, over different \textit{IoU} threshold ranges including the commonly used range of [0.5:0.95]. As shown in Fig. \ref{fig:AP_iou2}b), the \textit{AP} scores on the commonly-used \textit{IoU} range substantially differ from those on lower \textit{IoU} ranges. Note that the corresponding AP for a \ti{point in mask} criterion would be 0.73. \\

\noindent\textbf{\textit{IoU} vs. \textit{Point in Mask}:} Considering the clinical goal of prioritizing the localization of polyps more than their boundaries, we compared the values of the aggregated metrics \textit{Sensitivity, Positive Predictive Value (PPV), F1-Score, F2-Score} and \textit{Average Precision} using point-based localization criteria to the values obtained  using \textit{Box IoU}. The result is shown in Tab.~\ref{tab:boxvspoint}.
 \textit{Point inside reference} criteria yield higher scores across all metrics compared to \textit{Box IoU} over most \textit{IoU} thresholds. This especially holds true for detection \textit{Sensitivity}. \\

\begin{table}[]
\begin{center}
\begin{tabular}{lcccc}
\toprule
\multirow{2}{*}{\begin{tabular}[c]{@{}l@{}}Localization \\ criterion\end{tabular}} & \multirow{2}{*}{\begin{tabular}[c]{@{}l@{}}Box \\ \ti{IoU}=0.5\end{tabular}} & \multirow{2}{*}{\begin{tabular}[c]{@{}l@{}}Point \\ inside Box\end{tabular}} & \multirow{2}{*}{\begin{tabular}[c]{@{}l@{}}Point \\ inside Mask\end{tabular}} & \multirow{2}{*}{\begin{tabular}[c]{@{}l@{}}Point \\ inside Hull\end{tabular}} \\
 &  &  &  &  \\
 \cmidrule(r){1-1} \cmidrule(r){2-2} \cmidrule(l){3-5}
\textit{Sensitivity} & 0.68 & 0.74 & 0.74 & 0.74 \\
\textit{PPV} & 0.79 & 0.86 & 0.85 & 0.86 \\
\textit{F1-Score} & 0.73 & 0.8 & 0.79 & 0.8 \\
\textit{F2-Score} & 0.70 & 0.74 & 0.74 & 0.74 \\
\textit{AP} & 0.66 & 0.73 & 0.73 & 0.73 \\
 \botrule
\end{tabular}
\end{center}
\caption{Point-based versus overlap-based localization criteria applied to the set of all six centers. Point-based criteria give rise to similar results while the \textit{Box Intersection over Union (IoU)} criterion consistently yields lower values.  }
\label{tab:boxvspoint}
\end{table}

\subsection{Reflection of domain interests}\label{sec:ReflectionDOM}
\begin{figure}[]
  \centering
  \includegraphics[width=11.5cm]{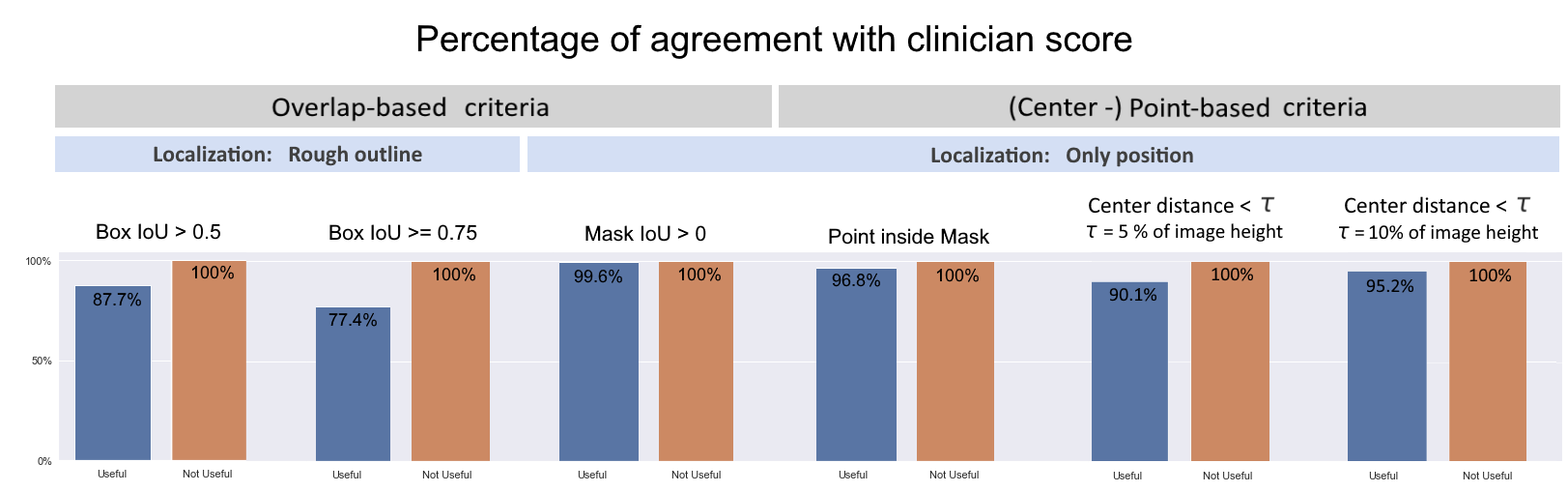}
\caption{Agreement of common localization criteria with clinicians’ ratings. Predictions rated as “not useful” by clinicians were rejected by all criteria without exception. However, especially overlap-based localization criteria yielded a high proportion of false negatives that clinicians would have classified as “useful”. Almost perfect agreement was achieved by the metric \textit{Mask IoU} $>$  0.}
\label{fig:rating_agreement3}
\end{figure}

In the presence of many sources of variability depending on the metric configuration, we conducted an experiment to determine which configuration aligns most with the clinical goal. We presented colonoscopy images of over 300 patients with their predicted bounding boxes to three gastroenterologists, one with over five years and two with over ten years of experience, who rated the predicted boxes as (clinically) “useful” or “not useful”. Each clinician was responsible for one third of the images and each image was only rated once. 
In order to assess the agreement of certain metric configurations with the clinician score, we plotted the number of predictions that met the criterion as a fraction of the predictions rated as “useful”, as well as the number of predictions not meeting the criterion as a fraction of predictions rated as “not useful”.
We applied overlap-based and point-based criteria and highlighted the localization granularity that they focus on (rough outline or only position). 
The result can be seen as a bar plot in Fig. \ref{fig:rating_agreement3}. All predictions clinically rated as “not useful” were rejected by all localization criteria. Criteria that focus only on position yielded a higher agreement with the "useful" score than those that localize based on overlap using rough outline.

\section{Discussion}\label{sec:discussion}
To our knowledge, we were the first to systematically investigate the variability of polyp detection performance resulting from various validation design choices. The following key insights can be derived from our experiments:
\ti{(1) Performance results are highly sensitive to various design choices:} Our experiments clearly demonstrate that various validation design choices have a substantial effect on the performance computed for object detection algorithms according to popular metrics. These range from the choice of test set to the specific metric configuration used. While the effect of using different classification metrics may be increasingly well-understood \cite{MaierHein2022Metrics}, we believe that common metrics, such as \ti{AP}, are often regarded as black boxes and the effect of the various hyperparameters remains poorly understood. Our findings clearly suggest that hyperparameters -- specifically the localization criterion and the corresponding threshold -- should not indiscriminately be adopted from other work, but carefully be chosen to match the domain need. \\
\ti{(2) Common metric configurations do not reflect the clinical need:} According to a usefulness assessment of polyp predictions from over 300 patients by three clinicians from different hospitals, commonly used localization criteria that are popular in the computer vision community do not reflect the clinical domain interest when deciding whether a prediction should be assigned a true positive or false positive. The community should therefore revisit the question of whether a good object detection method must necessarily yield a good outline of a polyp. Restricting the need to just localizing a polyp via its position (reflected by the requirement of $IoU > 0$, for example) might better approximate the clinical need and at the same time overcome problems resulting from suboptimal \textit{IoU} thresholds.\\
\ti{(3) Common hyperparameters may be too restrictive:} Our visual examples (Fig. \ref{fig:overlap_scheme3}) demonstrate that even fairly well-localized polyps feature an \textit{IoU} below the commonly used threshold of 0.5, resulting in them being considered a miss even though a clinician might find the prediction useful. The community may therefore want to reconsider commonly used threshold ranges and use a broader range (see Fig. \ref{fig:AP_iou2}b)).\\
\ti{(4) Comparison of performance across datasets can be largely misleading:} Our work finds that detection performance depends crucially on the polyp sizes. Hence, even if the prevalences of polyps across centers are similar, comparison of algorithm results can be largely misleading in case of different polyp size distributions.

The closest work to ours was recently presented by Ismail et al. \cite{ismail2021metrics} outside the field of deep learning. They provide anecdotal evidence on the non-comparability of confusion matrices between different methods, but do not analyze common multi-threshold metrics such as \ti{AP} or popular localization criteria that serve as the basis for popular classification metrics. Other related work focused on providing benchmarking data sets~\cite{fitting2022video} or showing limitations of metrics for clinical use cases outside the field of polyp detection~\cite{Kofler2021AreWU,Reinke2021CommonLO,Gooding2018ComparativeEO}.

A limitation of our study can be seen in the fact that we only used one object detection model. As a consequence, we are restricted to bounding boxes as predicted instances. On the other hand, the applied model was the winner of a very recent polyp detection challenge and can therefore be regarded as representative of the state of the art. Furthermore, almost all common object detection algorithms are based on predicting bounding boxes. Another limitation could be seen in the fact that we reported our findings only on a single data set~\cite{ali2021polypgen}. However, this data set comprises images from six centers and can therefore be seen as sufficiently representative for the scope of our research question. Finally, there are several other factors related to performance assessment that we did not prioritize in this work. These include the assignment strategy, the prevalence as well the confidence threshold in the case of counting metrics. Future work could hence explore the impact of these factors. 

In conclusion, our study is the first to systematically demonstrate the sensitivity of commonly used performance metrics in deep learning-based colon cancer screening to a range of validation design choices. In showing clear evidence for the disparity between commonly used metric configurations and clinical needs, we hope to raise awareness for the importance of adapting validation in machine learning to clinical relevance in general, and spark the careful reconsideration of common validation strategies in automatic cancer screening applications in particular.

\section*{Declarations}\label{sec:Declarations}

\noindent\textbf{Funding}
This project was supported by a Twinning Grant of the German Cancer Research Center (DKFZ) and the Robert Bosch Center for Tumor Diseases (RBCT). \\
\textbf{Competing interests}
The authors have no relevant financial or non-financial interests to disclose.\\
\textbf{Ethics approval}
This work was conducted using public datasets of human
subject data made available by \cite{Ali2022EndoCV21}.\\
\textbf{Consent to participate}
Not applicable.\\
\textbf{Consent for publication}
Not applicable.\\
\textbf{Availability of data and materials}
Not applicable.\\
\textbf{Code availability}
Not applicable.\\
\textbf{Authors' contributions}
All authors contributed and commented on previous versions of the manuscript. All authors read and approved the final manuscript.


\newcommand{\doi}[1]{\url{https://doi.org/#1}}
\newcommand{\ao}[1]{Accessed #1}

\end{document}